%% file: root.tex
\documentclass[letterpaper, 10 pt, conference]{ieeeconf} 

\IEEEoverridecommandlockouts                              
\usepackage{times}
\usepackage{epsfig}
\usepackage{graphicx}
\usepackage{grffile}
\usepackage{amsmath}
\usepackage{amssymb}
\usepackage{cite}

\input{modules/0.preamble.tex}

\usepackage[pagebackref=true,breaklinks=true,letterpaper=true,colorlinks,bookmarks=false]{hyperref}

\begin{document}

\input{modules/1.head.tex}

\maketitle

\input{modules/2.abstract.tex}

\input{modules/3.introduction.tex}

\input{modules/4.related_work.tex}

\input{modules/5.method.tex}

\input{modules/6.experiment.tex}

\input{modules/7.discussion.tex}

\bibliographystyle{IEEEtran}
\bibliography{egbib}

\end{document}

%% file: modules/0.preamble.tex
\usepackage{comment}
\usepackage{booktabs, multirow} 
\usepackage{soul}
\usepackage[table]{xcolor} 
\usepackage{placeins}

\usepackage[skip=1ex,labelfont=bf, textfont=it]{caption}
\usepackage[textfont=rm]{subcaption}

%% file: modules/1.head.tex
\title{\LARGE \bf
Optical Flow boosts Unsupervised Localization and Segmentation}


\author{Xinyu Zhang$^{1}$ and  Abdeslam Boularias$^{2}$
\thanks{$^{1}$Xinyu Zhang, Department of Computer Science, Rutgers University, 
        {\tt\small xz653@rutgers.edu}}%
\thanks{$^{2}$Abdeslam Boularias, Department of Computer Science, Rutgers University,
        {\tt\small ab1544@cs.rutgers.edu}}%
}



%% file: modules/2.abstract.tex
\begin{abstract}

Unsupervised localization and segmentation are long-standing robot vision challenges that describe the critical ability for an autonomous robot to learn to decompose images into individual objects without labeled data. These tasks are important because of the limited availability of dense image manual annotation and the promising vision of adapting to an evolving set of object categories in lifelong learning. Most recent methods focus on using visual appearance continuity as object cues by spatially clustering features obtained from self-supervised vision transformers (ViT). In this work, we leverage motion cues, inspired by the common fate principle that pixels that share similar movements tend to belong to the same object. 
We propose a new loss term formulation that uses optical flow in unlabeled videos to encourage self-supervised ViT features to become closer to each other if their corresponding spatial locations share similar movements, and vice versa. 
We use the proposed loss function to fine-tune vision transformers  that were originally trained on static images. 
Our fine-tuning procedure outperforms state-of-the-art techniques for unsupervised semantic segmentation through linear probing,  without the use of any labeled data. This procedure also demonstrates increased performance over original ViT networks across unsupervised object localization and semantic segmentation benchmarks. Our code is available at \url{https://github.com/mlzxy/flowdino}.

\end{abstract}

%% file: modules/3.introduction.tex
\section{Introduction}

 The ability to localize and recognize objects in images is crucial for intelligent robots to effectively operate in the real world~\cite{spelke2007core}. 
A feature representation that can distinguish and localize different semantic entities in a given image is important for building downstream algorithms, and for making the robot's behavior naturally more interpretable by humans~\cite{kulkarni2019unsupervised}.
Dense prediction tasks, such as detection and segmentation, are downstream tasks that are particularly important in robotics. 
The current prevailing approach to these tasks is to train deep neural networks with large amounts of human-labeled dense image annotations.
Despite the development of self-supervised learning~\cite{he2022masked}, and the utilization of vast amounts of Internet images and descriptions~\cite{radford2021learning}, densely annotated datasets are still scarce and obtained through manual annotation, which is not only a labor-intensive and expensive process, but also constrained to a fixed set of object categories without the ability to discover new objects. Therefore, unsupervised learning of dense prediction tasks without labeled data, including detection and segmentation, is an important open research challenge.

Current methods for unsupervised dense image understanding generally involve discovering basic structures such as foreground masks~\cite{nguyen2019deepusps, qin2019basnet, chen2019unsupervised}, contours~\cite{arbelaez2010contour}, or invariant mappings under transformations~\cite{ji2019invariant}. These structures then guide the learning of pixel-level embeddings, enabling the spatial differentiation of different objects~\cite{ji2019invariant, van2021maskcontrast, ke2022multiview, van2022discovering}. A new paradigm of unsupervised dense prediction has emerged in the past year to leverage self-distilled vision transformers (ViTs)~\cite{caron2021emerging}. Deep-Spectral\cite{melas2022deepspectral} and STEGO~\cite{hamilton2022stego} have achieved state-of-the-art segmentation results through spectral clustering and self-training on top of features extracted from frozen ViTs. Leopart~\cite{ziegler2022leopart} incorporates spatial feature clustering into the training of ViTs. These pioneering works apply the principle of visual appearance continuity as object cues on self-supervised features, pushing the boundaries of unsupervised image understanding to complex images.

In this paper, we draw inspiration from the common fate principle~\cite{koffka2013principles}, which posits that pixels tend to belong to the same object if they move in the same direction at the same speed, i.e., if they have the same optical flow. Optical flow has been extensively used for video object segmentation and tracking for its ability to easily capture moving objects~\cite{zhou2020matnet, liu2021emergence, yang2021motiongroup, guesswhatmove}. 
However, few attempts have been made to transfer the motion information in optical flow to the task of localizing and segmenting objects in still images. 
Rather than using optical flow for object tracking in videos, our approach is to mimic the human ability to observe objects moving patterns and learn transferable objects concept  for static images.

We follow the assumption studied in CrossPixel~\cite{mahendran2019cross}: \textit{pixels sharing similar motion are likely to belong to the same object and vice versa}. We refine this assumption by only considering pixels within a local neighborhood, removing background motion, and concentrating learning on regions with substantial movements. 
Our approach utilizes optical flow as an auxiliary regularization for ViT features in self-supervised learning to encourage the ViT network to produce similar features in locations that exhibit similar pixel motions.
Specifically, we first estimate optical flows from adjacent frames in existing unlabeled raw video datasets~\cite{wang2021unidentified, miao2021vspw, xu2018youtube} using off-the-shelf optical flow model~\cite{raft2020teed}. We then split feature and optical flow maps into local patches. For each patch, we minimize the KL divergence between the feature cosine similarity and the flow similarity measured with a customized RBF kernel. We use this flow-based loss function to fine-tune the vision transformers proposed in DINO~\cite{caron2021emerging}. We evaluate the features from the fine-tuned networks in two downstream tasks: (1) the latest unsupervised object localization procedure proposed in~\cite{melas2022deepspectral}, and (2) the unsupervised semantic segmentation evaluation protocols proposed in~\cite{ziegler2022leopart, van2021maskcontrast}. We demonstrate increased performance over the original ViT networks across these unsupervised dense vision prediction tasks. Our proposed approach outperforms the state-of-the-art in unsupervised semantic segmentation on Pascal VOC 2012 and Cocostuff-27 through linear probing, while preserving the discriminative power on ImageNet. We summarize our contributions as follows:

\begin{enumerate}
    \item  A new unsupervised fine-tuning procedure using optical flow that leverages the correlation between motion and objectness to encourage self-supervised ViT features to become closer if their corresponding spatial locations share similar flows in a local vicinity.
    
    \item Implementation and evaluation of the proposed procedure in the DINO self-supervised framework~\cite{caron2021emerging}. We demonstrate 
    increased performance over original networks across unsupervised object localization and semantic segmentation tasks, and outperform state-of-the-art techniques in unsupervised semantic segmentation through linear probing.
\end{enumerate}

%% file: modules/4.related_work.tex
\section{Related work}

\noindent\textbf{Self-supervised Learning.} 
Learning self-supervised visual representations has drawn significant interest in 
computer vision. Early work was based on learning through solving pretext tasks such as inpainting, 
jigsaw puzzles, and colorization~\cite{zhang2016colorful, pathak2016context, noroozi2016unsupervised}. Recent major successes are broadly based on
momentum-based contrastive learning~\cite{he2020momentum, chen2020improved, caron2021emerging}, and 
masked auto-encoding~\cite{he2022masked}, and natural language supervision~\cite{radford2021learning}. In particular, DINO~\cite{caron2021emerging} showed that self-supervised features obtained through vision transformer (ViT) architectures and self-distillation explicitly contain 
scene layout and boundary information.
These emerging properties inspired pioneering work on unsupervised object discovery and semantic segmentation based on existing
self-supervised ViT models \cite{van2021maskcontrast, van2022discovering, melas2022deepspectral, hamilton2022stego, ziegler2022leopart, tokencut2022}. While existing works employ pretrained ViT models as a sub-component of larger systems, our approach provides a loss function that can be seamlessly incorporated within existing self-supervised frameworks.



\vspace{1em}

\noindent \textbf{Optical Flow.} Optical flow is defined as the per-pixel motion between adjacent video frames. This concept was introduced to describe the visual stimulus of moving objects~\cite{lucaskanade1981iterative}. Based on the object continuity property, optical flow has drawn constant interest in video object segmentation~\cite{zhou2020matnet, liu2021emergence, yang2021motiongroup, guesswhatmove}.
With the recent advances in deep learning~\cite{lecun2015deep}, and the success in learning  optical flow from 
 synthetic datasets~\cite{flyingchairs2015dosovitskiy, flyingthings2016mayer}, off-the-shelf dense optical flow estimation networks have now become easily available~\cite{raft2020teed}. For example, CMP~\cite{zhan2019condmotionprop} predicts optical flow as a pretext task for self-supervised learning, and 
CrossPixel~\cite{mahendran2019cross} embeds pixels to match similarity of corresponding flow vectors. While these early work emphasize learning general representations from motion, few attempts have been made to leverage optical flow in learning for unsupervised object localization and semantic segmentation of still images.

\vspace{1em}

\noindent \textbf{Unsupervised Object Localization.} Object localization refers to the prediction of  bounding boxes of foreground objects in images. Early unsupervised techniques for learning object localization focused on mining co-occurring patterns amongst image collections~\cite{vo2021large, vo2020toward, cho2015unsupervised, vo2021large}.
Recent work explores mining examples from single images~\cite{collins2018deep, zhang2020object}. Significantly increased performance has been achieved with graph-based partitioning procedures that use pre-trained ViT architectures~\cite{simeoni2021lost, tokencut2022, melas2022deepspectral}. In this work, we build upon existing localization procedures, and show the significant benefit of fine-tuning ViT features with optical flow.

\vspace{1em}

\noindent \textbf{Unsupervised Semantic Segmentation.} Unsupervised semantic segmentation involves generating pixel-level prediction that can be closely mapped to semantic labels through clustering or linear projection. 
A popular paradigm is to extract structures from images, including foreground masks~\cite{nguyen2019deepusps, qin2019basnet, chen2019unsupervised},  contours~\cite{arbelaez2010contour} or invariant mapping under transformation~\cite{ji2019invariant}, then use the structures to guide the learning of pixel-level embeddings~\cite{ji2019invariant, van2021maskcontrast, ke2022multiview, van2022discovering}. Another rising approach originates from discovering emergent object information from DINO ViTs~\cite{caron2021emerging}. For example, both Deep-Spectral~\cite{melas2022deepspectral} and STEGO~\cite{hamilton2022stego} achieve impressive segmentation results using features of frozen ViTs with spectral clustering and self-training. Leopart~\cite{ziegler2022leopart} achieves significant improvement by leveraging visual appearance continuity and incorporating cluster assignment loss~\cite{caron2020unsupervised} into DINO's self-supervised framework.
Our method utilizes a loss fomulation to finetune pretrained ViTs with motion cues from optical flow. We increase the performance on unsupervised semantic segmentation while preserving the discriminative power of original ViTs, and our loss formulation does not depend on DINO's self-supervised framework.

\vspace{1em}

\noindent \textbf{Vision Transformers.} Transformer architectures are the key behind the recent significant  success in natural language processing~\cite{vaswani2017attention}. Vision transformers (ViT) employ positional embedding and self-attention layers instead of convolutional layers~\cite{dosovitskiy2020image}.
Recent variants of ViT have demonstrated various advantages over traditional CNN architectures, including
higher computational efficiency~\cite{liu2021swin}, improved self-supervised learning efficacy~\cite{chen2020improved, caron2021emerging},
stronger performance on downstream vision tasks~\cite{carion2020detr, xie2021segformer}, and more interpretable features~\cite{raghu2021vit_vs_cnn}.
In this work, our technique is applied to ViTs.

\begin{figure*}[t!]
\begin{center}
\includegraphics[width=\linewidth]{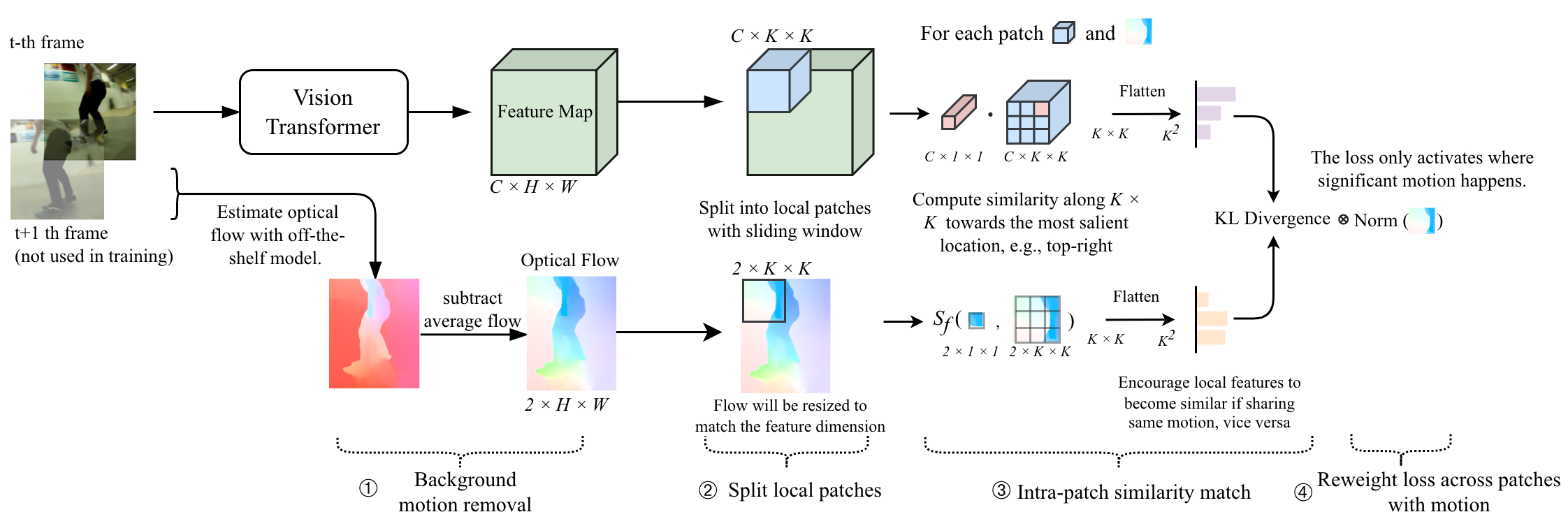}
\end{center}
   \caption{\textnormal{Workflow of our proposed optical flow loss}. 
   Features and flows are divided into patches, and their similarity matrices are matched with KL divergence, which are then averaged with flow norm to focus learning on moving areas. 
   Optical flows are computed from adjacent video frames and subtracted with average to reduce background motion.
   }
\label{fig:main}
\end{figure*}

%% file: modules/5.method.tex
\section{Method}

\subsection{Motivation}

Our approach stems from a simple assumption: \textit{Pixels that share similar motion (i.e., optical flow) are likely to belong to the same object, and vice versa.} While this assumption has been thoroughly studied in previous works such as CrossPixel~\cite{mahendran2019cross}, there are several limitations in those works:

\begin{enumerate}
    \item Two pixels having similar motion may not belong to the same object if they are far apart. 
    
    \item For pixels with static motion, it cannot be determined whether they belong to the same object or not. 
    
    \item Camera-motion may induce a similar motion at every pixel, including foreground and background pixels.
\end{enumerate}

To address the limitations, we refine the assumption with the following constraints 
that impose a reliable correlation between motion and appearance.
\begin{enumerate}
    \item Pixels sharing a similar motion are presumed to belong to the same object only if they are within the same local neighborhood.
    
    \item Objectness learning is focused on pixels with substantial motion. That is, learn only from regions where significant motion actually occurs.
    
    \item Optical flow is normalized to reduce the effect of background motion, which results from camera-motion.
\end{enumerate}

The locality assumption goes back to the earlier works of Lucas and Kanade~\cite{lucaskanade1981iterative}, which assumed that optical flow remains constant within a local neighborhood, and can be represented by flows at a small number of interest points determined by visual features. 
 We focus here on a local vicinity and only learn from pixels with substantial flow.
In our proposed approach, detailed in the following section, we train a ViT to return similar visual features for pixels that have similar motion. 

\subsection{Approach}

We start by extracting optical flow from a video using an off-the-shelf model, and normalizing the optical flow to remove the effect of background motion. Next, we divide both the feature map and the flow map into local patches, and compute for each patch a loss that encourages feature similarity among pixels that share similar motions.
Lastly, we calculate a weighted average of patch-level losses with the local flow norm serving as the patch's weight to ensure that we only learn from patches with significant motion. In the following, we denote optical flow as 
$\mathbf{v} \in \mathbb{R}^{2\times H \times W}$, and feature map as $\mathbf{f} \in \mathbb{R}^{C\times H \times W}$. The overall workflow of the proposed approach is illustrated in Figure~\ref{fig:main}.


Given that optical flow can only be estimated from a pair of adjacent video frames, it cannot be obtained from standard image-based datasets. To overcome this challenge, we rely on unlabeled videos to augment existing standard image-based datasets.
The dataset preparation procedure is detailed in Section~\ref{sec:exp}.

\vspace{1em}

\begin{figure}
\setlength\belowcaptionskip{-15pt}
\centering
\setkeys{Gin}{width=\linewidth}
\hfil
\begin{subfigure}{0.3\linewidth}
    \caption*{Image}
    \includegraphics{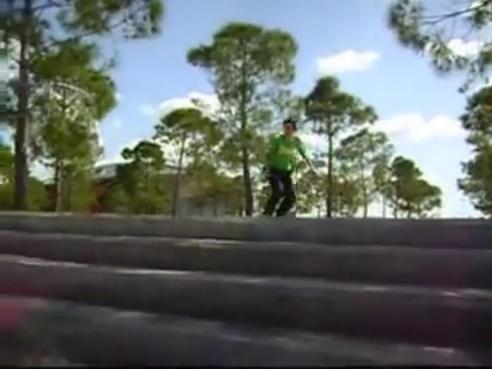}\\[2pt]
    \includegraphics{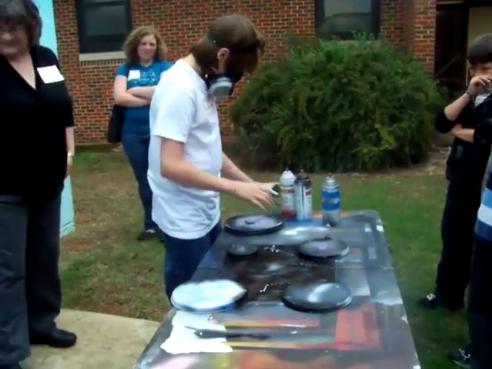}
    \end{subfigure}
\hfil
    \begin{subfigure}{0.3\linewidth}
    \caption*{Before}
    \includegraphics{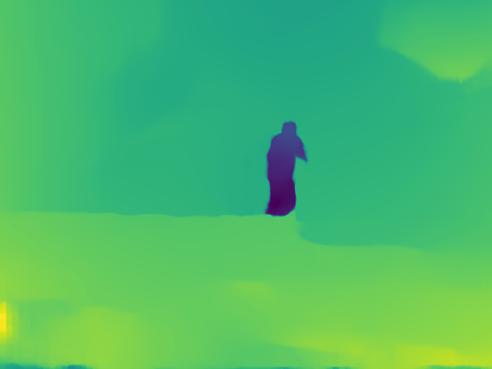}\\[2pt]
    \includegraphics{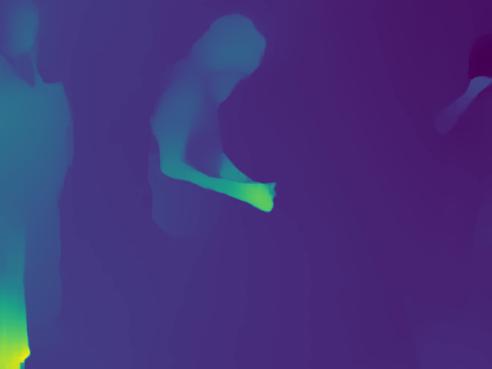}
    \end{subfigure}
\hfil
    \begin{subfigure}{0.3\linewidth}
    \caption*{After}
    \includegraphics{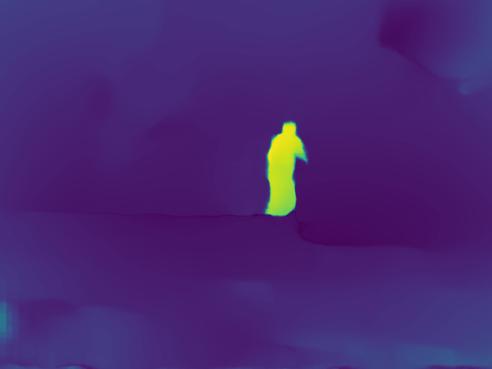}\\[2pt]
    \includegraphics{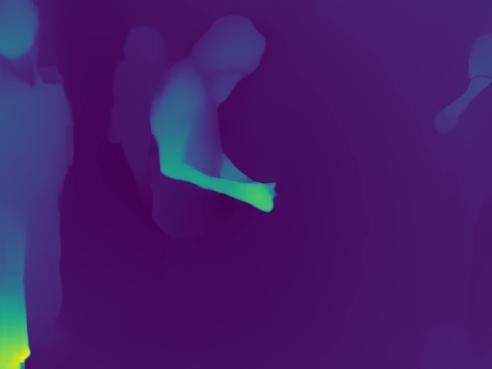}
    \end{subfigure}
\hfil
\caption{Comparing before vs. after background motion removal. From left to right, we show the original image, l2 norm of flows before and after motion removal.}
\label{fig:motion_removal}
\end{figure}

\noindent \textbf{Background motion removal.} 
We empirically found that simply subtracting the average flow significantly reduces the background motion. We further normalize the flow and project it to $[-1, 1]$ by dividing it by the maximum norm, as shown in Equation~\ref{eq:bmotion_removal}. We denote the stabilized and normalized flow as $\tilde{\mathbf{v}}$. 
We discuss the limitation of this approach in Section~\ref{sec:discussion}. Examples are shown in Figure~\ref{fig:motion_removal}.

\begin{equation}
\tilde{\mathbf{v}} = \frac{\mathbf{v} - \overline{\mathbf{v}}}{\parallel \mathbf{v} - \overline{\mathbf{v}} \parallel_\infty}   
 \label{eq:bmotion_removal}
\end{equation}

\vspace{0.5em}

\noindent \textbf{Split local patches.} We divide feature map $\mathbf{f}$ and flow $\tilde{\mathbf{v}}$ into patches with sliding windows. We denote the feature patches as $\mathbf{f}_p \in \mathbb{R}^{C\times K \times K}$ and flow patches as  $\mathbf{v}_p \in \mathbb{R}^{2\times K \times K}, p \in \{1,...,L\}$, where $K$ denotes patch size and $L$ denotes the number of patches.

\vspace{0.8em}

\noindent \textbf{Intra-patch similarity match.} For each patch, we calculate $K\times K$ similarity matrices for both feature and optical flow by measuring the similarity between the most salient pixel and other $K\times K$ locations. The most salient pixel is selected as the one with the strongest self-attention received at class token.
Specifically, we flatten and denote the feature and flow similarity matrices to be vectors $\mathbf{z}_{f,p}, \mathbf{z}_{v,p} \in \mathbb{R}^{K^2}$. We denote by $\mathbf{f}_{p,i} \in \mathbb{R}^C$ and $\mathbf{v}_{p,i} \in \mathbb{R}^2$, where $i \in \{1,..,K^2\}$, the feature and flow vectors at location $i$ within patch $p$. We denote the most salient location in patch $p$ as $s_p$. 
The feature similarity vector $\mathbf{z}_{f,p}$ is computed with the cosine similarity, i.e., the dot product of features after normalizing them to a unit length, as shown in Equation~\ref{eq:f_z}.

\vspace{-0.4em}

\begin{equation}
\mathbf{z}_{f,p} = [..., \mathbf{f}_{p, s_p} \cdot \mathbf{f}_{p,i},...]^\top\quad i \in \{1, ..., K^2\}
\label{eq:f_z}
\end{equation}

The flow similarity vector $\mathbf{z}_{v,p}$ is computed with a customized RBF kernel function $S_f$, as shown in Equation~\ref{eq:v_z}.
\begin{equation}
\mathbf{z}_{v,p} = [..., S_f(\mathbf{v}_{p, s_p}, \mathbf{v}_{p,i}),...]^\top \quad i \in \{1, ..., K^2\}
\label{eq:v_z}
\end{equation}
The RBF kernel function $S_f$ is given in Equation~\ref{eq:Sf}, where $cos(x, y)$ denotes cosine similarity with the output saturated to $[0, 1]$, and $\sigma$ is the RBF's radius parameter. The exponential term is multiplied by $\parallel\mathbf{y}\parallel_2$ to separate stationary pixels, which form clear boundaries between foreground and background. 

\begin{equation}
S_f (\mathbf{x}, \mathbf{y}) = \parallel \mathbf{y} \parallel_2 \exp ((\cos(\mathbf{x} , \mathbf{y}) - 1)  / \sigma )
\label{eq:Sf}
\end{equation}

Next, we transform the feature and flow similarity to a probability distribution using softmax as shown in Equation~\ref{eq:softmax}, where $\tau$ is a temperature parameter. We denote the flow and feature distributions as  $\mathbf{p}_{v,p}$ and $\mathbf{p}_{f,p}$, respectively.

\begin{equation}
\mathbf{p}_{\cdot,p} = \text{softmax}( \mathbf{z}_{\cdot, p} / \tau)
\label{eq:softmax}
\end{equation}

Then, we minimize the KL divergence between $\mathbf{p}_{v,p}$ and $\mathbf{p}_{f,p}$ in Equation~\ref{eq:Lp} to encourage features of pixels with similar motions to become closer and vice versa.
We denote the KL divergence loss of patch $p$ as $\mathcal{L}_p$, given as
\begin{equation}
\mathcal{L}_p = D_{KL}(\mathbf{p}_{v,p} \parallel \mathbf{p}_{f,p}).
\label{eq:Lp}
\end{equation}

\noindent \textbf{Reweighting loss terms across patches with motion.} To concentrate the learning on areas with significant motion, the weight $w_p$ of patch $p$ is the proportion of patch $p$'s motion relative to the motions of all the patches in the given frame, as outlined in Equation~\ref{eq:w_p}.

\vspace{-0.5em}

\begin{equation}
w_p = \frac{\parallel \mathbf{v}_p \parallel_2}{\sum_{p=1}^L \parallel \mathbf{v}_p \parallel_2}.
\label{eq:w_p}
\end{equation}

Finally, the overall loss is given as a weighted average of the local patch losses, as described in Equation~\ref{eq:L}. 

\vspace{-0.5em}

\begin{equation}
\mathcal{L} =  \sum_{p=1}^L w_{p} \mathcal{L}_p.
\label{eq:L}
\end{equation}

\vspace{-0.5em}

We use this optical flow loss in combination with the original self-supervised learning loss to train our vision transformer network, as depicted in Figure~\ref{fig:main}.

%% file: modules/6.experiment.tex
\section{Experimental results}
\label{sec:exp}

To evaluate the effectiveness of the proposed training procedure, we perform experiments on both unsupervised object localization and unsupervised semantic segmentation using the features before and after motion-guided fine-tuning. We implement this procedure using PyTorch~\cite{paszke2019pytorch} on top of the released implementation of DINO~\cite{caron2021emerging}, and apply it on ViT-Small and ViT-Base~\cite{dosovitskiy2020image} architectures with patch sizes of 8 and 16 respectively, and with weights initialized from the pre-trained DINO models~\cite{caron2021emerging}. We set temperature $\tau$ to 0.1 and radius $\sigma$ to 0.7 in all experiments.
During fine-tuning, each batch consists of half of the images taken from ImageNet~\cite{deng2009imagenet}, and the other half is made of video frames. Our final loss is the sum of DINO's loss on the ImageNet images and the optical flow loss on the video frames.

\vspace{1em}

\noindent\textbf{Dataset Preparation.} To create datasets  with optical flow information, we merge the following existing video datasets: UVO~\cite{wang2021unidentified}, VSPW~\cite{miao2021vspw}, and Youtube-VOS~\cite{xu2018youtube}. We extract frames from about 10,000 videos with a frame interval of five and estimate the optical flow between each adjacent frames with the Raft-Large model~\cite{raft2020teed}. We also apply the same procedure on the Moment-in-Time dataset~\cite{monfortmoments}, but we use only the first dataset except when otherwise noted. Compared to images, each pixel in the optical flow is stored in two 32-bit float numbers rather than a single 8-bit unsigned integer. Optical flow does not have a widely available compression format. Therefore, storing the optical flow requires roughly one to two orders of magnitude more space than an equivalent number of images. To overcome this challenge, we quantize the 
 optical flow into 16-bit integers and concatenate the flow in the x and y directions into a single 32-bit float number.
Next, we save the 32-bit stream in the TIFF image format, which supports 32-bit float pixel values, and we apply the TIFF compression protocol. By applying this approach to store the optical flow, we observe about a ten-times reduction in disk space usage.

\subsection{Object Localization}

We extract features with our motion fine-tuned ViT models and apply the latest unsupervised object localization method from Deep-Spectral~\cite{melas2022deepspectral}. 
As is standard practice, we compare the results to prior work on three datasets: Pascal VOC 2007, Pascal VOC 2012~\cite{everingham2009pascal}, and COCO-20k~\cite{vo2020toward} (a subset of 20K images from MS-COCO dataset~\cite{lin2014microsoft}). 
We follow the evaluation procedure used in~\cite{vo2020toward, simeoni2021lost}, which accepts one bounding box for each image. 
Results are reported in the Correct Localization (CorLoc) metric, which measures the percentage of images on which one object can be correctly localized by the given bounding box. An object is considered to be correctly localized if the predicted bounding box has a greater than 50\% intersection-over-union (IoU) with the object's ground-truth bounding box.

\begin{table}[h]\centering
\setlength\belowcaptionskip{-10pt}
\begin{tabular}{lccc}\toprule
\textbf{Method} &\textbf{VOC-07} &\textbf{VOC-12} &\textbf{COCO-20k} \\\midrule
Selective Search~\cite{uijlings2013selective} &18.8 &20.9 &16 \\
EdgeBoxes~\cite{zitnick2014edge} &31.1 &31.6 &28.8 \\
Kim et al.~\cite{kim2009unsupervised} &43.9 &46.4 &35.1 \\
Zhang et al.~\cite{zhang2020object} &46.2 &50.5 &34.8 \\
DDT+~\cite{wei2019unsupervised} &50.2 &53.1 &38.2 \\
rOSD~\cite{vo2020toward} &54.5 &55.3 &48.5 \\
LOD~\cite{vo2021large} &53.6 &55.1 &48.5 \\
DINO-[CLS]~\cite{caron2021emerging} &45.8 &46.2 &42.1 \\
LOST~\cite{simeoni2021lost} &61.9 &64 &50.7 \\
Deep-Spectral~\cite{melas2022deepspectral} &62.7 &66.4 &52.2 \\
Deep-Spectral* &60.5 &65.7 &48.5 \\
Ours (Deep-Spectral) &\textbf{63.1(+2.6)} &\textbf{68.5(+2.8)} &\textbf{53.6(+5.1)} \\
\bottomrule
\end{tabular}
\caption{Single-object localization performance (CorLoc). Our results are obtained by reusing the post-processing procedures in Deep-Spectral~\cite{melas2022deepspectral} with our motion fine-tuned features, without any supervision. `*' denotes the results we reproduce from the official released implementations.}\label{tab:obj_loc}
\end{table}

\begin{figure}[h]
\centering
\setkeys{Gin}{width=\linewidth}
\begin{subfigure}{0.23\linewidth}
    \caption*{Image}
    \includegraphics{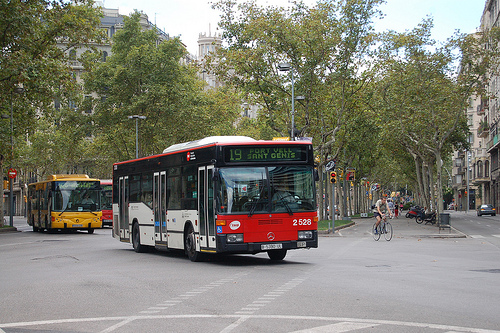}\\[1pt]
    \includegraphics{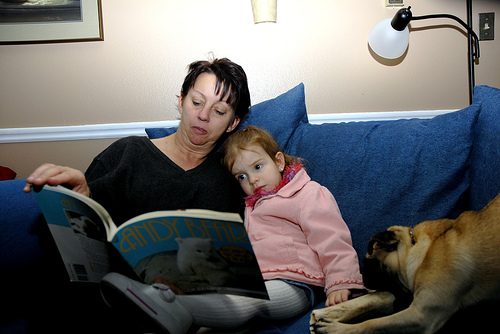} \\[1pt]
    \includegraphics{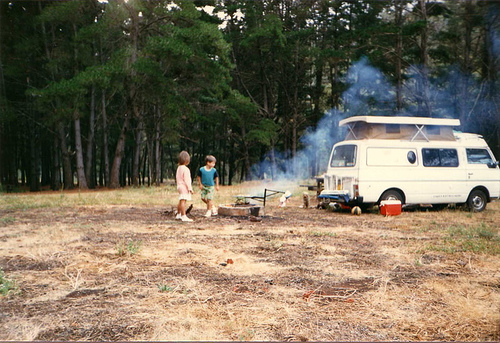}
    \end{subfigure}
    \begin{subfigure}{0.23\linewidth}
    \caption*{Baseline}
    \includegraphics{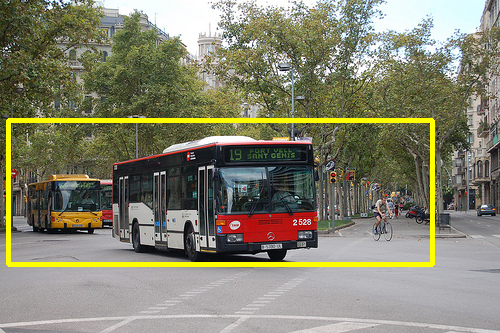}\\[1pt]
    \includegraphics{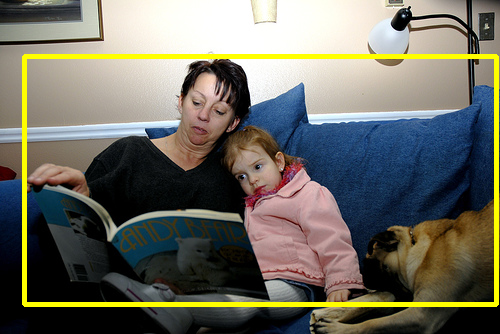}\\[1pt]
    \includegraphics{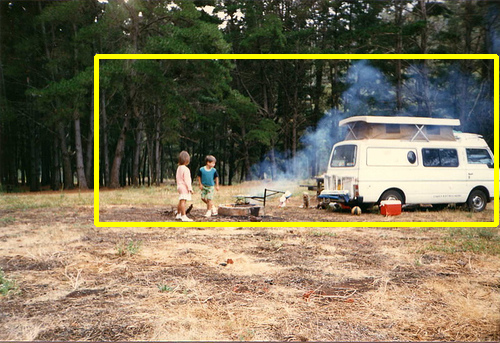}
    \end{subfigure}
    \begin{subfigure}{0.23\linewidth}
    \caption*{Ours}
    \includegraphics{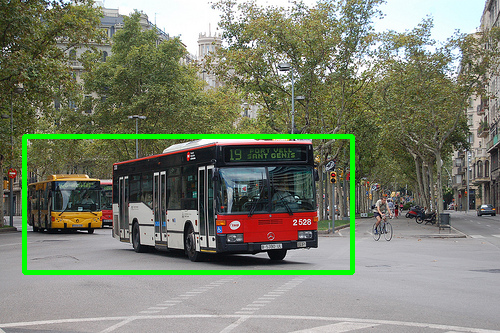}\\[1pt]
    \includegraphics{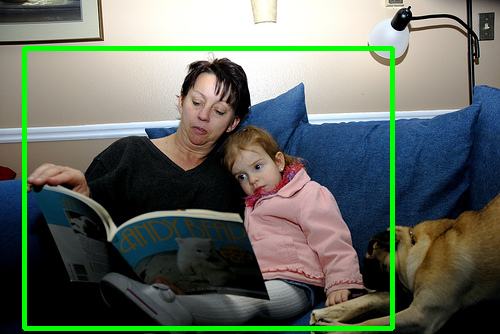}\\[1pt]
    \includegraphics{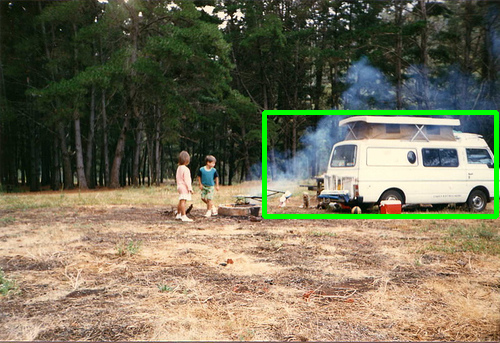}
    \end{subfigure}
    \begin{subfigure}{0.23\linewidth}
    \caption*{Ground Truth}
    \includegraphics{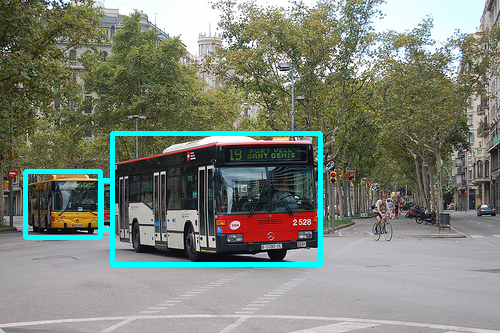}\\[1pt]
    \includegraphics{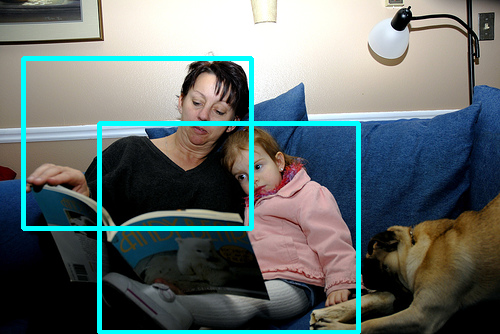}\\[1pt]
    \includegraphics{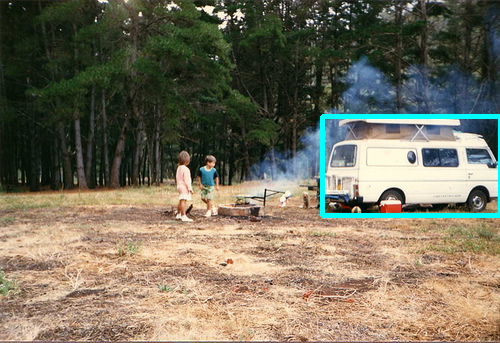}
    \end{subfigure}
\hfil
\caption{Object localization on Pascal VOC 2012. From left to right, we show the original image, predicted bounding boxes from Deep-Spectral~\cite{melas2022deepspectral}, our predicted bounding boxes using the same procedure, and ground-truth boxes.}
\label{fig:obj_loc}
\end{figure}

The quantitative results are summarized in Table~\ref{tab:obj_loc}. 
We reproduce the localization performance in Deep-Spectral~\cite{melas2022deepspectral} using ViT-B8 (ViT-Base network with patch size 8) and the released implementation and hyper-parameters. We report our results by reusing the same implementation on the ViT-B8 architecture with our fine-tuned weights. Clear and consistent improvement over the original features can be seen in all three datasets. Note that this improvement is obtained automatically, without any human effort, because our approach is fully self-supervised. In Figure~\ref{fig:obj_loc}, we show some qualitative examples of our methods.

\begin{table}[h]\centering
\begin{tabular}{lrrrrr}\toprule
&\textbf{Network} &\textbf{VOC-07} &\textbf{VOC-12} &\textbf{COCO-20k} \\\midrule
\multirow{4}{*}{Baseline} &ViT-S16 &57.4 &63.4 &46.4 \\
&ViT-B16 &56.7 &62.8 &46 \\
&ViT-S8 &59.4 &64.1 &47.6 \\
&ViT-B8 &60.5 &65.7 &48.5 \\ \midrule
\multirow{4}{*}{Ours} &ViT-S16 &\textbf{58.4} &\textbf{64.3} &\textbf{48.1} \\
&ViT-B16 &\textbf{59} &\textbf{64.1} &\textbf{48.5} \\
&ViT-S8 &58.4 &\textbf{64.2} &46.7 \\
&ViT-B8 &\textbf{63.1} &\textbf{68.5} &\textbf{53.6} \\
\bottomrule
\end{tabular}
\caption{Comparing, in terms of object localization performance with Deep-Spectral algorithm~\cite{melas2022deepspectral}, different ViT architectures before vs. after fine-tuning them with motion. Motion-driven fine-tuning generally yields better results.}\label{tab:obj_loc_comp}
\end{table}
\begin{table}[h]\centering
\footnotesize
\begin{tabular}{lcccc}\toprule
\textbf{Arch} &\textbf{\scriptsize Motion Removal} &\textbf{\scriptsize  Patch Size} &\textbf{\scriptsize Loss Reweight} &\textbf{VOC-12} \\\midrule
ViT-S16 &\checkmark &5 &\checkmark &62.8 \\
ViT-S16 &\checkmark &3 &\checkmark &64.3 \\
ViT-S16 & &3 &\checkmark &62.8 \\
ViT-S16 & &3 & &59.85 \\ \midrule
ViT-S16 &\multicolumn{3}{l}{Baseline} &63.4 \\
ViT-S16 &\multicolumn{3}{l}{Only add video frames} &62.6 \\
ViT-S16 &\multicolumn{3}{l}{add video frames + flow loss} & 64.3 \\
\midrule
ViT-B8 &\multicolumn{3}{l}{Baseline} &65.7 \\ 
ViT-B8 &\multicolumn{3}{l}{Only add video frames} &64.88 \\
ViT-B8 &\multicolumn{3}{l}{add video frames + flow loss} & 68.5 \\
\bottomrule
\end{tabular}
\caption{Comparing different options of motion-driven fine-tuning on the object localization task (CorLoc).}\label{tab:main_ablation}
\end{table}

The results in Table~\ref{tab:obj_loc_comp} demonstrate that our motion-driven fine-tuning approach improves object localization performance across different ViT architectures. Additionally, in Table~\ref{tab:main_ablation}, we conduct an ablation study to assess the impact of our optical flow loss. Specifically, we compare our proposed optical flow loss with using only the DINO's self-supervised loss while adding video frames as extra training data. Our results reveal that solely adding video frames as new training images does not lead to performance gains. Moreover, we assess the individual contributions of background motion removal and motion-based loss re-weighting and demonstrate that performance deteriorates when background motion is not removed or when the learning is not focused on regions with substantial motion.

\subsection{Semantic Segmentation}

We evaluate our motion fine-tuned ViT models on unsupervised semantic segmentation. Our evaluation protocol follows prior work in self-supervised learning~\cite{ziegler2022leopart}, i.e., linear probing and cluster probing. Linear probing protocol involves training an extra linear projection from model outputs to ground-truth labels with supervision while freezing the model weights. Cluster probing protocol involves dividing spatial features into separate groups with clustering algorithms and applying Hungarian matching \cite{kuhn1955hungarian} to match clusters to ground-truth labels optimally. Both linear and cluster probing are done solely for the purpose of evaluating the learned features. 
We compare our results to prior methods on Cocostuff-27~\cite{caesar2018cvpr} and Pascal VOC 2012. Results are measured in terms of mean intersection-over-union (mIoU), which  denotes the percentage of overlaps between the predicted segmentation mask and the ground-truth across different classes.

\begin{table}[!htp]\centering
\small
\begin{minipage}{0.49\linewidth}
\begin{tabular}{lc}\toprule
\textbf{Method } &\textbf{\footnotesize Cocostuff-27} \\\midrule
ResNet50\cite{he2016deep} &10.2 \\
MoCoV2\cite{chen2020improved} &13.2 \\
MDC\cite{cho2021picie} &13.3 \\
PiCIE\cite{cho2021picie} &13.9 \\
PiCIE+H\cite{cho2021picie} &14.8 \\
STEGO\cite{hamilton2022stego} &41.2 \\
DINO\cite{caron2021emerging} &42.2 \\
Leopart\cite{ziegler2022leopart} &44.1 \\\midrule
Ours &\textbf{46.1(+2.0)} \\
\bottomrule
\end{tabular}
\end{minipage}\hfill
\begin{minipage}{0.49\linewidth}
\begin{tabular}{lc}
\toprule
\textbf{Method} &\textbf{VOC-12} \\\midrule
IIC\cite{ji2019invariant} &28 \\
MoCoV2\cite{chen2020improved} &45 \\
InfoMin\cite{tian2020makes} &45.2 \\
SWAV\cite{caron2020clustercontrast} &50.7 \\
SegSort\cite{hwang2019segsort} &36.2 \\
Hierach. Group.\cite{zhang2020self}\hspace{-2em} &48.8 \\
MaskContrast\cite{van2021maskcontrast}\hspace{-2em} &63.9 \\
Leopart~\cite{ziegler2022leopart} &68 \\\midrule
Ours &\textbf{68.7(+0.7)} \\
\bottomrule
\end{tabular}

\end{minipage}
\caption{Semantic Segmentation (mIoU) on Cocostuff-27 and Pascal VOC 2012 with linear probing. Our results are obtained using the evaluation protocol of Leopart~\cite{ziegler2022leopart}.}
\label{tab:uss_linear}
\end{table}

\begin{figure}[!ht]
\setlength\belowcaptionskip{-8pt}
\centering
\setkeys{Gin}{width=\linewidth}
    \begin{subfigure}{0.23\linewidth}
    \caption*{Image}
    \includegraphics{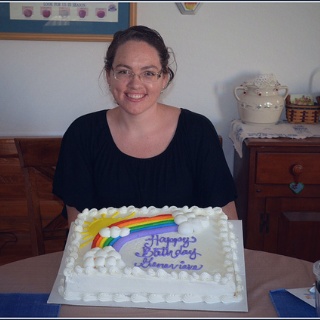}\\[1pt] 
    \includegraphics{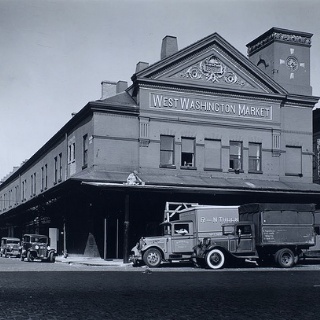}
    \end{subfigure}
    \begin{subfigure}{0.23\linewidth}
    \caption*{Baseline}
    \includegraphics{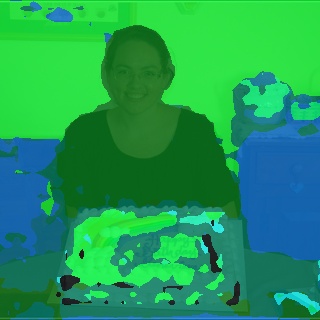}\\[1pt]
    \includegraphics{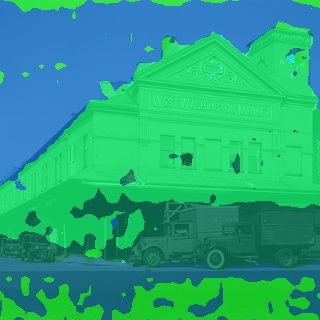}
    \end{subfigure}
    \begin{subfigure}{0.23\linewidth}
    \caption*{Ours}
    \includegraphics{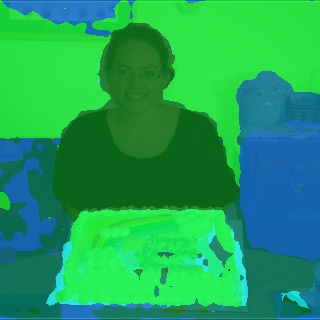}\\[1pt]
    \includegraphics{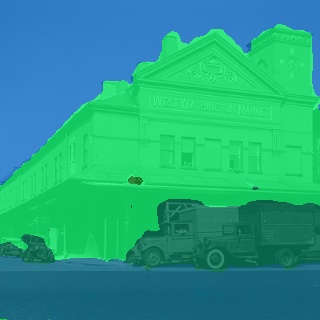}
    \end{subfigure}
    \begin{subfigure}{0.23\linewidth}
    \caption*{Ground Truth}
    \includegraphics{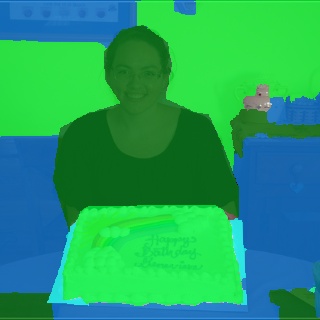}\\[1pt]
    \includegraphics{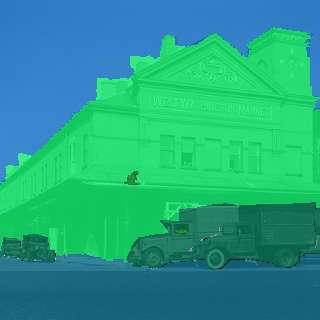}
    \end{subfigure}
\caption{Semantic Segmentation on Cocostuff-27 through linear probing. From left to right, we show the original image, predicted bounding boxes from ViT before and after motion-driven fine-tuning, and ground-truth semantic masks.}
\label{fig:uss_cocostuff}
\end{figure}

Table~\ref{tab:uss_linear} compares our approach with previous methods through linear probing, including the state-of-the-art technique Leopart~\cite{ziegler2022leopart}. Our results are obtained by fine-tuning on ViT-B8 with the Moment-in-Time dataset~\cite{monfortmoments} without ImageNet. It should be noted that while the result reported by Leopart~\cite{ziegler2022leopart} is achieved using a smaller ViT architecture (ViT-S16), our method demonstrates the potential to improve semantic segmentation performance solely by utilizing motion information, without relying on existing visual continuity through spatial clustering. Qualitative examples of our approach are shown in Figure~\ref{fig:uss_cocostuff}.

\begin{table}[!htp]\centering
\begin{tabular}{lrr}\toprule
\textbf{Method} &\textbf{mIoU} \\\midrule
Co-Occurrence\cite{isola2015learning} &4 \\
CMP\cite{zhan2019condmotionprop} &4.3 \\
Colorization\cite{zhang2016colorful} &4.9 \\ 
IIC\cite{ji2019invariant} &9.8 \\
MaskContrast\cite{van2021maskcontrast} &35 \\
Deep-Spectral\cite{melas2022deepspectral} (w/o self-training) &30.8 ± 2.7 \\
Deep-Spectral\cite{melas2022deepspectral} &37.2 ± 3.8 \\ \midrule
\textit{DINO Baselines} & \\ \midrule
ViT-B16 &27.9 ± 1.18 \\
ViT-S16 &30.2 ± 1.15 \\ \midrule
\textit{Ours (w/o self-training)} \\ \midrule
ViT-B16 &31.0 ± 1.6 (\textbf{+3.1}) \\
ViT-S16 &35.35 ± 2.2 (\textbf{+5.15}) \\
\bottomrule
\end{tabular}
\caption{Semantic Segmentation (mIoU) on Pascal VOC 2012 through cluster probing. We adopt the evaluation protocol from MaskContrast~\cite{van2021maskcontrast} and use the same ViT from Deep-Spectral~\cite{melas2022deepspectral} without the self-training part of~\cite{melas2022deepspectral}.
}\label{tab:uss_clst}
\end{table}

\begin{figure}[!ht]
\centering
\setkeys{Gin}{width=\linewidth}
    \begin{subfigure}{0.23\linewidth}
    \caption*{Image}
    \includegraphics{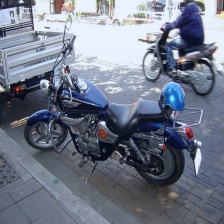}\\[1pt] 
    \includegraphics{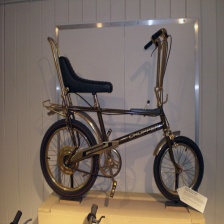}
    \end{subfigure}
    \begin{subfigure}{0.23\linewidth}
    \caption*{Baseline}
    \includegraphics{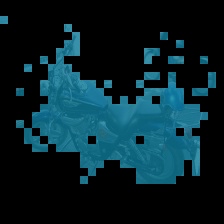}\\[1pt]
    \includegraphics{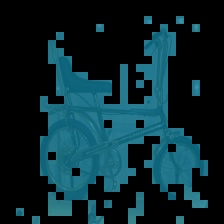}
    \end{subfigure}
    \begin{subfigure}{0.23\linewidth}
    \caption*{Ours}
    \includegraphics{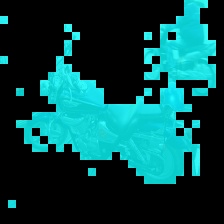}\\[1pt]
    \includegraphics{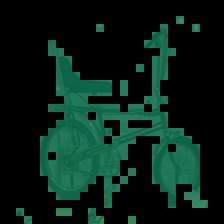}
    \end{subfigure}
    \begin{subfigure}{0.23\linewidth}
    \caption*{Ground Truth}
    \includegraphics{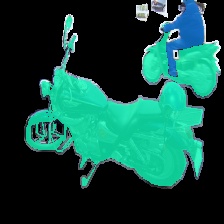}\\[1pt]
    \includegraphics{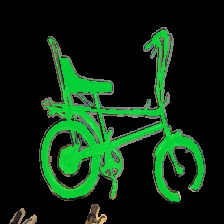}
    \end{subfigure}
\caption{Semantic Segmentation on Pascal VOC 2012 through cluster probing. From left to right, we show the original image, predicted bounding boxes from ViT before and after motion-driven fine-tuning, and ground-truth semantic masks.}
\label{fig:uss_voc}
\end{figure}

In Table~\ref{tab:uss_clst}, we compare to previous methods through cluster probing on Pascal VOC 2012. Due to the lack of a  standardized practice on cluster probing, we adopt the evaluation protocol from MaskContrast~\cite{van2021maskcontrast}, but use the ViT attention as estimated saliency instead of a supervised saliency network. We select other existing works based on similarity of their evaluation protocols. 
Notably, our approach outperforms the DINO baselines and achieves comparable results to Deep-Spectral without self-training. It is worth mentioning that self-training is a general performance boosting technique that involves training on pseudo-labels generated for the target dataset. 
Qualitative examples are shown in Figure~\ref{fig:uss_voc}.

\begin{table}[!htp]\centering
\begin{tabular}{lrrrrr}\toprule
& &\textbf{VOC-12} &\textbf{Cocostuff-27} &\textbf{ImageNet} \\\midrule
\multirow{2}{*}{Baseline} &ViT-S16 &47.41 &36.1 &74.44 \\
&ViT-S8 &49.53 &38.6 &78.33 \\ \midrule
Leopart~\cite{ziegler2022leopart} &ViT-S16 &68 &44.1 &51.99 \\\midrule
\multirow{2}{*}{Ours} &ViT-S16 &59.39 &39.96 &73 \\
&ViT-S8 &63.28 &41.26 &77.46 \\
\bottomrule
\end{tabular}
\caption{Comparing networks before vs. after motion-driven fine-tuning on unsupervised semantic segmentation (mIoU), and ImageNet classification (top-1 accuracy) tasks.}\label{tab:uss_lc_imagenet}
\end{table}

\begin{table}[!htp]\centering
\setlength\belowcaptionskip{-10pt}
\begin{tabular}{lcccc}\toprule
&ViT-S16 &ViT-B16 &ViT-S8 &ViT-B8 \\\midrule
Baseline &74.44 &75.87 &78.33 &77.28 \\
Fine-tuned &73.6 &74.65 &77.45 &76.59 \\
\bottomrule
\end{tabular}
\caption{Comparing before vs. after fine-tuning using video frames but without our optical flow loss on ImageNet classification (top-1 accuracy). 
}\label{tab:imagenet_without_flow_loss}
\end{table}

Table~\ref{tab:uss_lc_imagenet} compares our approach with the baselines DINO and Leopart~\cite{ziegler2022leopart} on linear probe segmentation and ImageNet classification performance. The ImageNet classification accuracy is evaluated using a weighted nearest neighbor classifier (k-NN) as in \cite{wu2018unsupervised}. It should be emphasized that our approach not only boosts semantic segmentation performance compared to the baselines, but also preserves the discriminative power on ImageNet. In contrast, the discriminative power is significantly affected in Leopart despite the remarkable improvement on segmentation tasks. Table~\ref{tab:imagenet_without_flow_loss} shows the ImageNet top-1 classification accuracy after fine-tuning with video frames without our optical flow loss. It can be observed that a similar amount of performance drop occurs even without using our optical flow loss. This further suggests that the optical flow loss may not be the main reason for the accuracy drop observed in Table~\ref{tab:uss_lc_imagenet}.

%% file: modules/7.discussion.tex
\section{Discussion and Conclusion}
\label{sec:discussion}

We have shown that improved object understanding can be achieved for certain self-supervised learners through learning from motion information that is embedded in adjacent video frames. This information is readily available from off-the shelf  optical flow estimators. 
Some open questions are, however, still worth discussing. 
The background motion removal through mean reduction is likely to leave extra background motions around the image corners due to camera distortion, and it may also  
 diminish small object movements. 
 Despite the generality of our procedure and its independence of DINO's loss, our current implementation and experiments are still closely linked to the self-supervision training of DINO. This suggests the potential for designing more general modules that could translate motion into objectness information, which can be agnostically digested by other networks.

As visual continuity and motion are both intrinsic clues for determining objectness, 
the possibility of unifying them in a single framework has yet to be fully explored, while the lack of quality video datasets like ImageNet will likely continue to be a limiting factor.
Beyond only leveraging adjacent frames, it is possible to extract long-term spatial-temporal correspondences from videos to further improve representation learning for still images.
